\documentclass{article}

\usepackage{arxiv}

\usepackage[utf8]{inputenc} 
\usepackage[T1]{fontenc}    
\usepackage{hyperref}       
\usepackage{url}            
\usepackage{booktabs}       
\usepackage{amsfonts}       
\usepackage{nicefrac}       
\usepackage{microtype}      
\usepackage{lipsum}
\usepackage{graphicx}
\usepackage{color}
\usepackage{tablefootnote}
\usepackage{multirow}
\usepackage{soul}
\usepackage{makecell}
\graphicspath{ {./images/} }

\title{Forecasting Crude Oil Price Using Event Extraction}

\author{
	Jiangwei Liu \\
	School of Information Management and Engineering\\
	Shanghai University of Finance and Economics\\
	Shanghai 200433, China \\

	\And
	Xiaohong Huang \\
	School of Information Management and Engineering\\
	Shanghai University of Finance and Economics\\
	Shanghai 200433, China \\

}

\begin{document}
\maketitle

\begin{abstract}
	Research on crude oil price forecasting has attracted tremendous attention from scholars and policymakers due to its significant effect on the global economy. Besides supply and demand, crude oil prices are largely influenced by various factors, such as economic development, financial markets, conflicts, wars, and political events. Most previous research treats crude oil price forecasting as a time series or econometric variable prediction problem. Although recently there have been researches considering the effects of real-time news events, most of these works mainly use raw news headlines or topic models to extract text features without profoundly exploring the event information. In this study, a novel crude oil price forecasting framework, AGESL, is proposed to deal with this problem. In our approach, an open domain event extraction algorithm is utilized to extract underlying related events, and a text sentiment analysis algorithm is used to extract sentiment from massive news. Then a deep neural network integrating the news event features, sentimental features, and historical price features is built to predict future crude oil prices. Empirical experiments are performed on West Texas Intermediate (WTI) crude oil price data, and the results show that our approach obtains superior performance compared with several benchmark methods.
\end{abstract}

 

\section{Introduction}
\label{sec:introduction}
Crude oil plays a significant role in the global economy, for nearly one-third of the world's energy consumption comes from it. Also, oscillations in oil prices significantly affect the economy of oil-exporting and oil-importing nations \cite{Abdollahi-3, YuDai-13}. Accurate oil price forecasting would help policymakers adopt proper policy and make appropriate decisions regarding energy resources. However, crude oil price prediction has been a challenging problem in forecasting research because oil prices are affected by many factors. Except for the fundamental market factors, such as supply, demand, and inventory, oil price fluctuation is strongly influenced by economic development, conflicts, wars, and breaking news \cite{WuWang-4}. For example, oil producers were paying buyers to take the commodity off their hands over fears that storage capacity could run out in May 2020, and WTI oil price even turned negative for the first time in history on 20 April 2020. Another recent example is that crude oil price movements have exhibited a stronger correlation with the severe degree of the COVID-19 pandemic \cite{AtriKouki-12, HuangZheng-48, Gil-AlanaMonge-46, LiuWang-47}. The challenge is characterizing and modeling such nonlinear and nonquantitative factors because most of such information is contained in raw texts.

Overall, existing research on crude oil price forecasting can be categorized into three main classes: econometric models (including time-series models), machine learning or deep learning methods, and hybrid approaches. Among time series approaches, Autoregressive Integrated Moving Average (ARIMA), Generalized Autoregressive Conditional (GARCH), Empirical Mode Decomposition (EMD), and Complete Ensemble Empirical Mode Decomposition (CEEMD) are primarily used \cite{AhmedShabri-15, WuChen-14, WuWu-16}. For example, \cite{Abdollahi-3} uses CEEMD to decompose the original oil prices into five nonlinear and three volatile components (IMFs). Then the author uses MS-GARCH to model and forecast volatile components and SVM-PSO (Support Vector Machine - Particle Swarm Optimization) to model and predict nonlinear components, respectively. Lastly, the linear addition of these forecasts gives a more reliable estimation of the oil price. Econometric models, especially structural models, focus on how specific economic factors and the behaviors of economic agents affect the future values of crude oil prices \cite{FreyManera-17}. For example, \cite{de-Albuquerquemellode-Medeiros-18} proposes a Self-Exciting Threshold Auto-regressive-SETAR model dealing with structural breaks in oil price longitudinal data, treatment, and forecasting.

Support vector machines (SVMs) and neural networks (NNs) are the most typical machine learning methods due to their extraordinary ability in modeling nonlinearity and volatility \cite{YuWang-20, YuZhang-19}. However, shallow architectures are insufficient to model complex patterns with numerous factors \cite{Bengio-21}. Recently, deep learning (DL) has become a mainstream approach in various fields. DL approaches explore complicated structures and patterns in large data sets using the backpropagation algorithm, which indicates how a machine should change its internal parameters \cite{LecunBengio-22}. Deep learning can represent highly nonlinear and highly varying functions; thus, DL-based approaches have also been widely used in oil price forecasting \cite{WuWang-4, WuWu-16}. 

Hybrid methods integrate the methods mentioned above and thus utilize their advantages synthetically. Reference \cite{Abdollahi-3, YuWang-20, AbdollahiEbrahimi-24, LinJiang-9} assemble time series decomposing models and AI models to enhance the forecasting performance. The reason why hybrid methods usually achieve better results than single models lies in two aspects. On the one hand, time-series approaches or econometric models specialize in capturing the linearity and volatility in price time series. On the other hand, AI models specialize in nonlinear and non-stationary characteristics. 

Recent research on utilizing news to enhance crude oil price forecasting can be divided into two categories. The first line of work mainly utilizes news sentiment \cite{LiXu-27, ZhaoZeng-26}, or Google trends \cite{WuWang-4} as auxiliary features. The second line of work concentrates on the news context. For example, \cite{BaiLi-28} uses the Latent Dirichlet Allocation (LDA) topic model to construct predictive factors; \cite{WuWang-4, WuWang-8} use Convolutional Neural Network (CNN) to extract text features from online news headlines automatically.

\begin{figure}[htbp]
	\centering
	\includegraphics[width= 6 in]{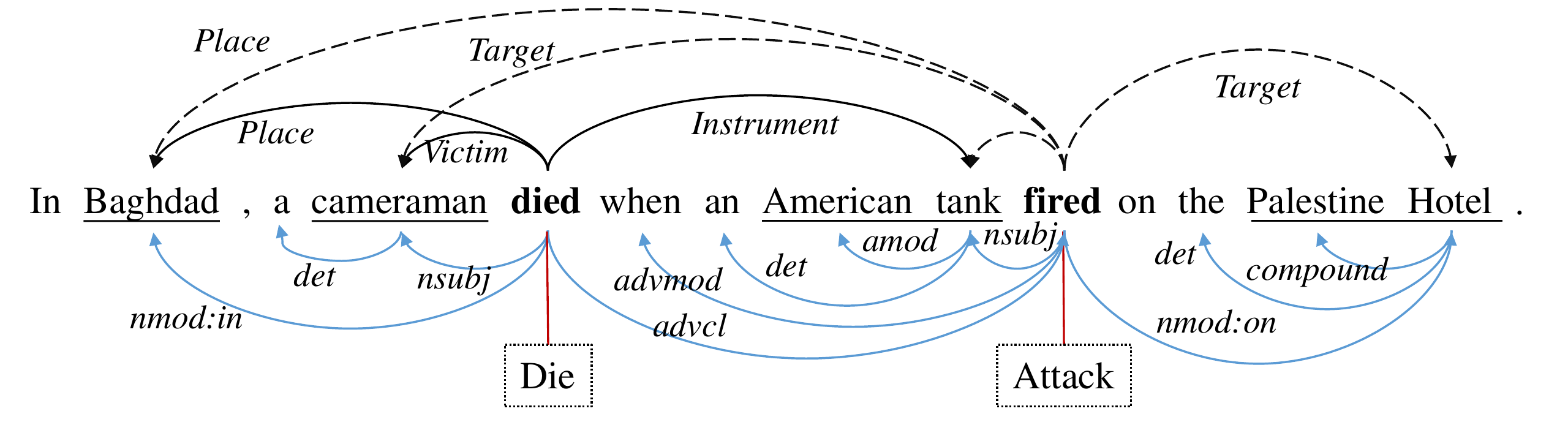}	
	\caption{An example of event extraction annotation and the syntactic parser results. There are two events in one sentence: Die and Attack. The upper arcs link event triggers to their corresponding arguments, with the argument roles on the arcs. The lower side demonstrates the syntactic parser results.}
	\label{fig1}	
\end{figure}

Event Extraction (EE) is an advanced form of Information Extraction (IE) that handles textual content or relations between entities, often performed after executing a series of initial NLP steps \cite{HogenboomFrasincar-53}. Event Extraction is different from the LDA topic models used in previous crude oil price forecasting literature. For example, there are two event types involved in sentence S1: "Die" and "Attack", triggered by "died" and "fired", respectively.
\begin{itemize}
	\item  \emph{S1: In Baghdad, a cameraman died when an American tank fired on the Palestine Hotel.}
\end{itemize}
For Die event, "Baghdad", "cameraman" and "American tank" are its arguments with corresponding roles Place, Victim, and Instrument, respectively. For Attach event, "Baghdad", "cameraman", "American tank" and "Palestine Hotel" are its arguments with corresponding role Place, Victim, Instrument and Target, respectively. This is a somewhat more complex example with three arguments shared, which is more challenging than the simple case with one event type in one sentence. Fig. \ref{fig1} shows an example of event extraction annotation and the syntactic parser results.

Considering that news messages reflect recent major events about the oil market, we argue that the oil market movements are not only sentimental-driven but also event-driven. And that event-extracted features are less noisy than raw headlines or news text. However, there is little literature on using event extraction to enhance oil price forecasting. To those ends, we utilize open domain event extraction algorithms to extract latent event types and event arguments as our event features. A neural network is also built to predict crude oil prices with historical prices, news sentiment, latent event types, and event arguments as input features. The three main contributions of this paper are presented as follows. (1) Open domain event extraction algorithms are introduced into crude oil price forecasting. (2) News sentiment is also utilized to enhance the prediction accuracy. (3) A novel framework AGESL that integrates multi-channel information (e.g., historical prices, latent event types, event arguments, and four sentimental factors) is proposed for forecasting crude oil price series.

The remainder of this paper is organized as follows. We first review related work and introduce the data sets and their preprocessing. Then we propose our framework and introduce its modules in detail. The experimental section reports results and discusses some characteristics of the proposed AGESL. Conclusions and future work are followed.

\section{Related Work}
\label{sec:related-work}
This section reviews existing typical literature on crude oil price forecasting and summarizes the literature in Table \ref{table1}. More details about the leading technologies used in the proposed framework AGESL are followed.

\begin{table}
	\caption{Summary of recent studies on crude oil price forecasting.}
	\label{table1}
	\centering
	\begin{tabular}{ p{60pt} p{150pt} p{200pt} }
		\toprule
		Literature& Features&  Models \\
		\midrule
		\cite{XieYu-5} (2006)&	Price&	SVM \\
		\cite{YuWang-20} (2008)&	Price&	EMD and Neural Network \\
		\cite{AhmedShabri-15} (2014)&	Price&	ARIMA, GARCH, SVM \\
		\cite{YuDai-13} (2016)&	Price&	Grid-GA-based Least squares support vector regression \\
		\cite{LiXu-27} (2017)&	Price and news text sentiment&	SVM, Decision Tree, and Back Propagation Neural Networks (BPNN) \\
		\cite{ZhaoLi-6} (2017)&	Price&	Stacked denoising autoencoders and bagging \\
		\cite{ElshendyColladon-41} (2018)& Price, sentiment,and text statitics & ARIMAX \\
		\cite{WuChen-14} (2019)&	Price&	CEEMD, ARIMA, and sparse Bayesian learning \\
		\cite{WuWu-16} (2019)&	Price&	EEMD and Long Short Term Memory (LSTM) \\
		\cite{ZhaoZeng-26} (2019)&	Price and web text sentiment&	Ridge regression, LASSO, support vector regression (SVR), BPNN, and random forest (RF) \\
		\cite{ZhaoLiu-40} (2019)& Price and news topics& A two-layer NMF (non-negative matrix factorization) model and GIHS (giant information history simulation) \\
		\cite{LiShang-42} (2019)& Price, sentiment, topics and market data& Convolutional Neural Network (CNN), Latent Dirichlet Allocation (LDA), Random Forest (RF), SVR and Linear Regression \\
		\cite{Abdollahi-3} (2020)&	Price&	CEEMD, SVM, PSO, and MS-GARCH \\
		\cite{LinJiang-9} (2020)&	Price&	WPD (wavelet packet de-noise)), EMD, and GARCH-M models \\
		\cite{SadikDate-7} (2020)&	Price and news sentiment&	Vector auto regression and Kalman (VAR) filtering framework \\
		\cite{BaiLi-28} (2021)&	Price, news topic, and sentiment&	LDA, SeaNMF, and AdaBoost RT \\
		\cite{WuWang-4} (2021)&	Price, news text, and Google trends&	CNN, BPNN, MLR (multiple linear regression), SVM, and GRU (gated recurrent unit) \\
		\cite{WuWang-8} (2021)&	Price, oil market data, and news headlines&	CNN, BPNN, SVM, RNN, and LSTM	\\
		\bottomrule
	\end{tabular}
	\label{tab1}
\end{table}

\subsection{ARIMA-GARCH}
Autoregressive Integrated Moving Average (ARIMA) and Generalized Autoregressive Conditional (GARCH) are two mainstream models in time series predicting. ARIMA was introduced by Box and Jenkins (1976), which is a linear combination of past values and past residuals. Bollerslev (1986) presented GARCH, which is a generalized form of the Autoregressive Conditional Heteroscedastic (ARCH) model initiated by Engle (1982) \cite{AhmedShabri-15}. The main difference is that ARIMA forecasts future values from past values and past residuals. Whereas GARCH focuses on the time-varying variance of residuals, also being called time-varying volatility. They can be separately used to model the time series or integrated to enhance the prediction performance.

Generally, a non-stationary time series can be transformed into a stationary one by differencing operation. Furthermore ARIMA is a typical method to model a stationary time series. A nonseasonal ARIMA model $ARIMA(p,d,q)$ has three integer parameters $p$, $d$, $q$: the order of AR, differencing, and MA, respectively. For example, a time series ${y_t}$ is said to be an $ARIMA(p,1,q)$ process if the transformation ${c_t}=\Delta {y_t}{\rm{ = }}{y_t}{\rm{-}}{y_{t-1}}=(1-L){y_t}$ follows a stationary and invertible $ARMA(p,q)$ model, where $L$ is a back-shift operator (also named lag operator). A general $ARMA(p,q)$ model is formulated as
\begin{equation} 
	{c_t} = c + \sum\limits_{i = 1}^p {{\alpha _i}} {c_{t - i}} + {\varepsilon _t} + \sum\limits_{{\rm{j}} = 1}^q {{\theta _j}{\varepsilon _{t - j}}} \label{eq1}
\end{equation}
where ${\varepsilon _t}$ is the white noise process with variance ${\sigma ^2}$, $c$ is intercept term, ${\alpha _i}$ and ${\theta _i}$ are coefficients of AR and MA processes.

Innovation series ${a_t}$ follows a $GARCH\left( {m,s} \right)$ model if 
\begin{equation} 
	{a_t} = {\sigma _t}{u _t} ,
\end{equation}
\begin{equation} 
	{\sigma _t}^2 = {\alpha _0} + \sum\limits_{i = 1}^{\rm{m}} {{\alpha _i}} a_{t - i}^2  + \sum\limits_{j = 1}^s {{\beta _j}\sigma _{t - j}^2} ,
\end{equation}
\begin{equation} 
	{u _t} \sim IID \left( {mean=0{\rm{ }}, variance=1} \right) ,
\end{equation}
where ${\alpha _0} > 0$, ${\alpha _{\rm{i}}} \ge 0$, ${\beta _{\rm{j}}} \ge 0$, and $\sum\nolimits_{{\rm{i}} = 1}^{\max (m,s)} {\left( {{\alpha _i} + {\beta _i}} \right)}  < 1$ .

GARCH models play an important role in financial risk evaluation scenarios. They are also beneficial for parameter estimation and thus help enhance prediction accuracy. Existing literature has demonstrated that the oil price innovation series usually exhibits volatility clustering and evolves. In our study, we use the GARCH model to capture the time-varying variance to adjust the amplitude of price movement.

\subsection{Sentiment Analysis}
Crude oil market is one of the most important commodity markets. Rich online information about crude oil market emerges every day, including massive news reports, which directly reﬂect diﬀerent driving factors and extreme events concerning oil price trends \cite{LiXu-27}. However, there are still limitations in exploring and using news text. Because directly utilizing raw oil news text may introduce noises or useless information. So how to effectively construct and employ high-quality features from news text becomes crucial. 

Recently, research on utilizing sentiment analysis to enhance oil price forecasting accuracy has become popular \cite{LiXu-27, ZhaoZeng-26, BaiLi-28, ElshendyColladon-41, LiShang-42, SadikDate-7}. Reference \cite{LiXu-27} uses Henry's Finance-Speciﬁc dictionary to filter the positive and negative words, then calculates the sentiment scores. Considering positive news tends to reduce volatility, whereas negative news tends to increase volatility, \cite{SadikDate-7} proposes a parametrized, nonlinear function to characterize the impact of news sentiment on price volatility. Reference \cite{ElshendyColladon-41} adopts two sentiment factors: a compound score and emotionality calculated as the deviation from neutral sentiment to facilitate crude oil price prediction, while \cite{LiShang-42} extracts two sentiment factors: a polarity score and a subjectivity score to facilitate crude oil price prediction. Reference \cite{ZhaoZeng-26} uses VADER (Valence Aware Dictionary and sEntiment Reasoner) method to analyze and predict the tendency of news text. VADER is a rule-based unsupervised method and has been used widely in text sentiment analysis. VADER is easy to use, and it outputs four sentiment scores: negative score, neutral score, positive score, and compound score when fed with one paragraph text. In our study, we use vaderSentiment, a Python implementation of VADER, to generate four corresponding sentiment factors.

\subsection{Event Extraction}
Another technology to construct and employ high-quality features from news text is Event Extraction that handles textual content or relations between entities \cite{HogenboomFrasincar-53}. The news reflects commercial, social, or unexpected political events that play important roles in the volatility of crude oil prices. Although there have been various pre-trained language models, such as Embeddings from Language Models (ELMo) \cite{PetersNeumann-31}, Generative Pre-Training model (GPT) \cite{RadfordNarasimhan-32}, Bidirectional Encoder Representations from Transformers (BERT) \cite{DevlinChang-33}, etc., and using word embeddings or sentence embeddings to quantify the semantic representation of the news is standard practice. Although it leads to a good performance in various applications, there are still deficiencies in two aspects. On the one hand, the embeddings could not always provide a convincing explanation. On the other hand, averaging the word embeddings may ignore some vital information, while sentence or paragraph embeddings may contain noises. Event Extraction technology can handle these challenges and answer 5W1H questions: who did what, when, where, why, and how about of an event. 

Existing event extraction approaches can be categorized into two main classes: domain-dependent event extraction and open domain event extraction. Domain-dependent event extraction focuses on extracting event arguments with event types and event schemas pre-specified. However, it is less useful for decision-making areas, such as security, financial forecasting, etc., which may require comprehensive knowledge on broad-coverage, fine-grained, and dynamically evolving event categories \cite{LiuHuang-34}. In comparison, open-domain event extraction could alleviate these challenges. Existing open domain event extraction approaches mainly utilize dependency parsing and rules to generate events \cite{ChauEsteves-36, Valenzuela-EscarcegaHahn-Powell-35}. Reference \cite{LiuHuang-34} extracted event type, schema, and arguments using a neural latent variable network and Bayesian inference model (ODEE) and got better results than other base models. 

\begin{figure}[htbp]
	\centering
	\includegraphics[width= 5 in]{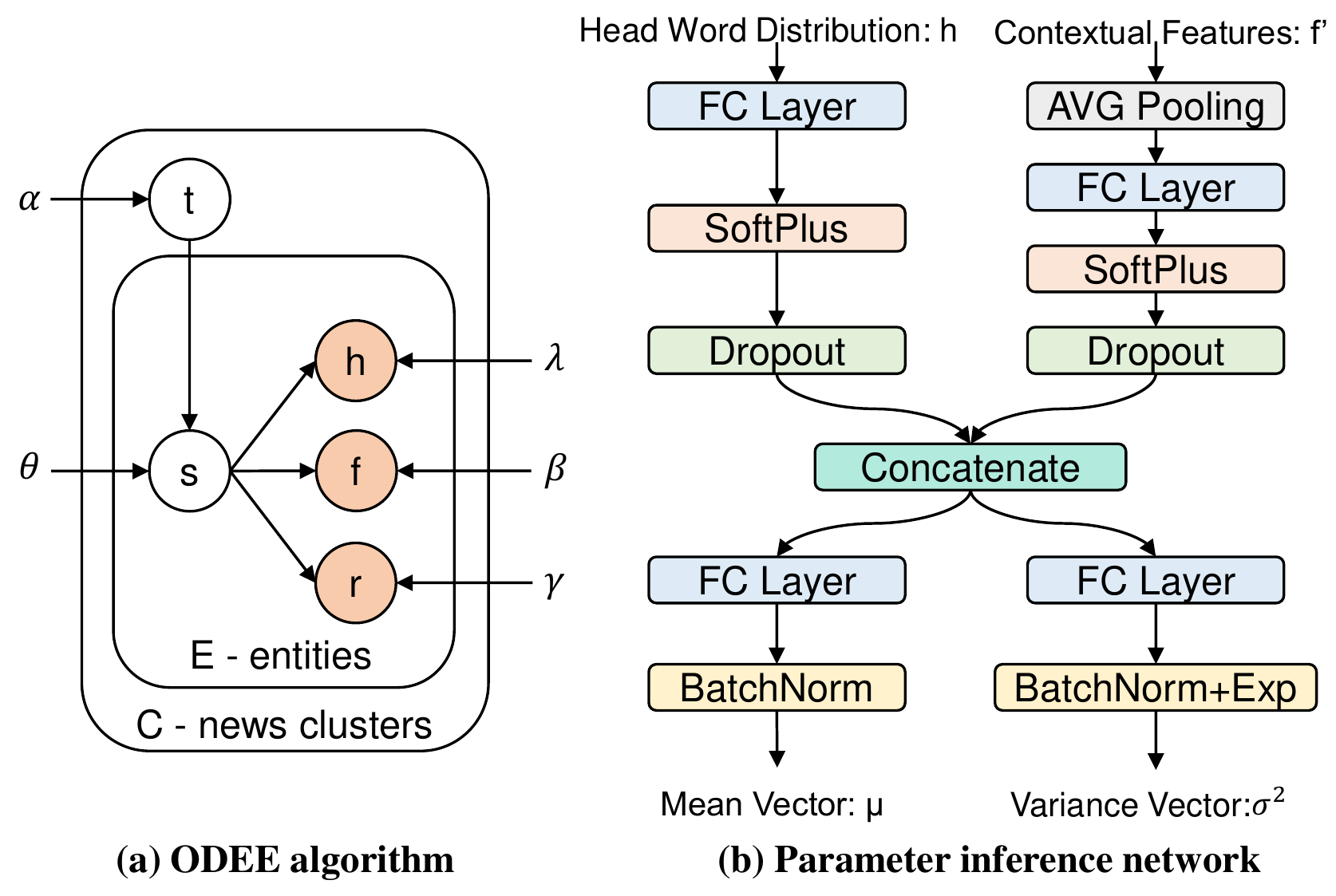}	
	\caption{ODEE algorithm and parameter inference network. $\alpha$ , $\theta$ ,$\lambda$ ,$\beta$ and $\gamma$ are parameters of sampling distributions. The related sampling distributions are as follows: a latent event type vector $t \sim Normal(\alpha)$, a slot $s \sim Multinomial(MLP(t; \theta))$, a head word $h \sim Multinomial(1, \lambda)$, a feature vector $f \sim Normal(\beta)$, and a redundancy ratio $r \sim Normal(\gamma)$.}
	\label{fig2}	
\end{figure}

In our study, we adopt the ODEE method to perform open domain event extraction. The main algorithm and the parameter inference network of ODEE are shown in Fig. \ref{fig2}.
In Fig. \ref{fig2}(a), for each news cluster of the given corpus, we first sample a latent event type from a normal distribution $Normal(\alpha)$; then for each entity, we sample a slot $s$ from a multinomial distribution  $Multinomial(MLP(t; \theta))$, a head $h$ from a multinomial distribution $Multinomial(1, \lambda)$, a feature vector $f$ from a normal distribution $Normal(\beta)$, and a redundancy ratio $r$ from a  normal distribution $Normal(\gamma)$. Fig. \ref{fig2}(b) is the inference network architecture to obtain the posterior distribution of the latent variables $s$ and $t$, given a news cluster $c$. More detailed information about the ODEE algorithm can be found in reference \cite{LiuHuang-34}.

\section{Data And Data Processing}
In this section, we introduce the crude oil price data and related news datasets and their preprocessing.

\subsection{Crude Oil Price Data}
\begin{figure}[htbp]
	\centering
	\includegraphics[width= 6 in]{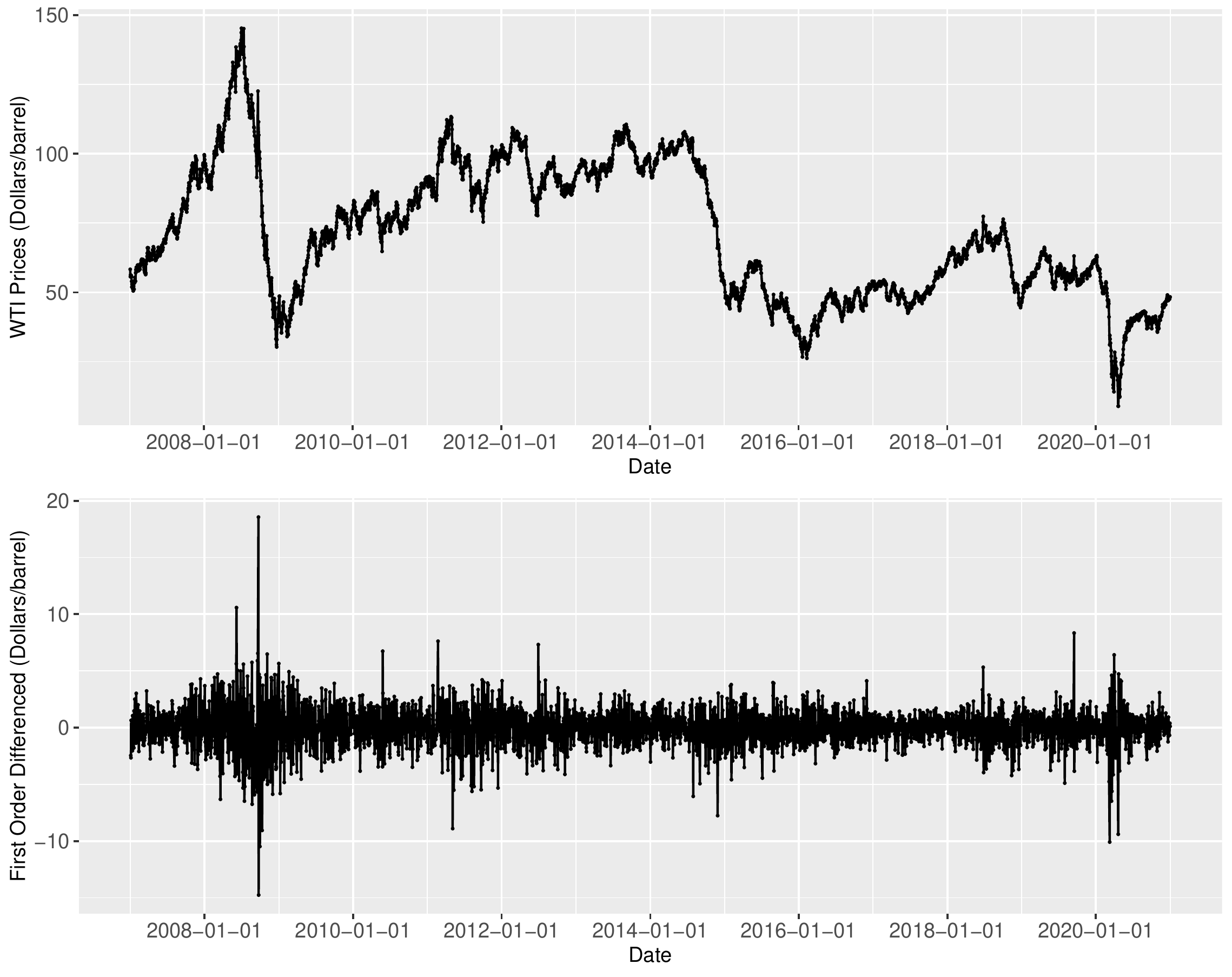}	
	\caption{Illustrations of WTI daily closing prices and first-order differenced series.}
	\label{fig3}	
\end{figure}

West Texas Intermediate (WTI) is one of the most critical global crude oil price indexes and can reflect global crude oil price movements in time. In this study, WTI daily prices are investigated. There are a total of 3522 observations ranging from January 2007 to December 2020. The original WTI series and its first difference series are illustrated in Fig. \ref{fig3}.

Consistent with previous literature \cite{WangZhou-2, ZhaoLi-6}, we split the data into training and testing in an 80/20 ratio. We hold 10\% training data as the validation set used to choose the optimal parameters in actual experiments. It is worth mentioning that the daily price of WTI turned negative for the first time in history on 20 April 2020 for various reasons. The key reason may be that oil producers were over fears that storage capacity could run out in May 2020. As shown in Fig. \ref{fig3}, this observation will be treated as an outlier, replaced by the lowest price nearby, and forecasted separately.

\subsection{News Data}
We also collect news data from The Guard\footnote{https://content.guardianapis.com/search, Accessed: 2021-07-21} and The New York Times\footnote{https://developer.nytimes.com/apis, Accessed: 2021-07-21} ranging from January 2007 to December 2020. Some measures should be taken to satisfy a particular requirement that the open domain event extraction method ODEE framework requires its inputs: clusters of related news. First, The Guard affords APIs to access its news headlines and contents, with a function that filter words can narrow the search down to news on crude oil. Second, we download the headlines, URLs, and leading paragraphs through The New York Times APIs, then crawl the news contents by the corresponding URLs. Third, the filter words, such as "crude oil" or "energy oil", are used to restrict the news being crude oil-related. After filtering, there are 17514 pieces of news from The Guard left, and that number for The New York Times is 13275. Finally, the whole news data is arranged grouping by date with an average of 8 news per day.

\section{Our Approach}
\begin{figure*}[htbp]
	\centering
	\includegraphics[width= 6.0 in]{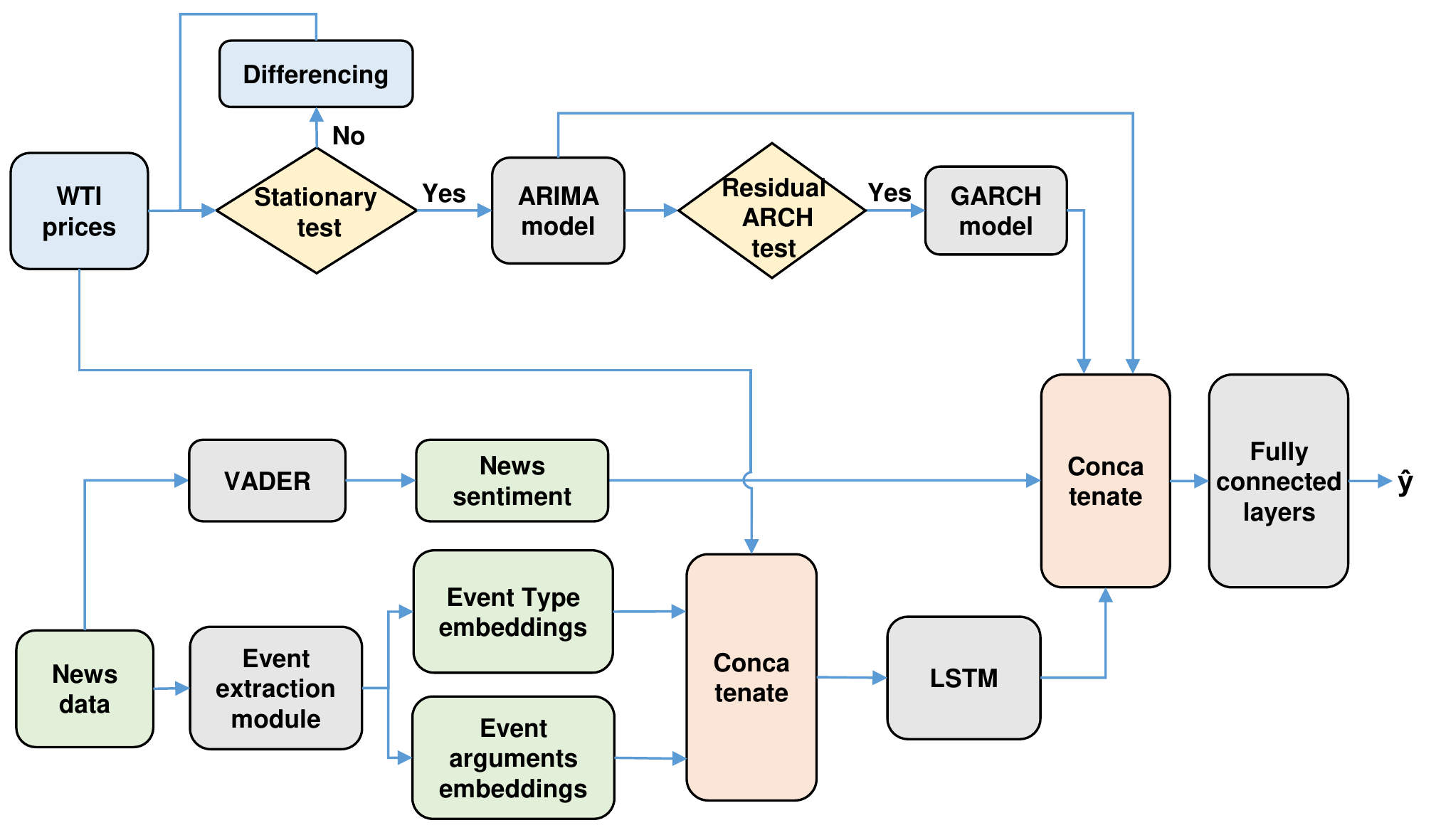}	
	\caption{Illustration of the proposed framework AGESL.}
	\label{fig4}	
\end{figure*}

This section introduces our crude oil price prediction framework AGESL, as illustrated in Fig. \ref{fig4}. Overall, AGESL contains five main modules: Mean price prediction (ARIMA model), Volatility prediction (GARCH model), Event extraction module, Sentiment analysis module, and LSTM module. Four kinds of features are used as input of AGESL, involving two data sources: historical prices and news text. Three features, historical prices excluded: sentiment scores, news types, and news arguments are extracted from news text. Mean price prediction and Volatility prediction modules are responsible for predicting future mean prices and fluctuation, mainly depending on historical price data or its varieties. The event extraction module contains a neural latent variable network and Bayesian inference model, with the responsibility to extract latent event type embeddings and event arguments. We concatenate event type embeddings, event arguments embeddings, and historical prices as input of the LSTM module. Then we concatenate the output of the ARIMA model, GARCH model, LSTM model, along with the formatted news sentiment sequence together as the input of the last fully connected layers. The last fully connected layers integrate and weigh the individual predictive model outputs, affording a synthetical prediction. The modules of AGESL are described step by step as follows.

\subsection{ Mean Price Prediction}
ARIMA module is responsible for predicting WTI future mean prices. Five steps are needed to fulfill this task. First, the Augmented Dicky-Fuller (ADF) test should be carried out to check the stationarity of the WTI daily price series. Second, differencing operation is used to generate a stationary time series. Third, the Autocorrelation function (ACF), Partial Autocorrelation function (PACF), and Information Criterion (AICC, AIC, and BIC) are incorporated to determine the AR and MA orders. Fourth, the Box-Pierce test is performed to check whether the residuals are independent white-noise, and if not, model parameters are readjusted. Lastly, the fitted model is used to forecast the future price rollingly.

\subsection{Volatility Prediction}
GARCH models can capture the main characteristics of the crude oil market, specifically leptokurtic distribution and volatility clustering. The volatility prediction task contains three main steps. First, Lagrange Multiplier (LM) test and Portmanteau Q test are performed to check the ARCH effect on the residuals of the ARIMA model. Second, GRARCH orders are determined by the Information Criterion and orders of the fitted ARIMA model. Lastly, the fitted GARCH model is used to forecast the volatility rollingly.

\subsection{Event Extraction}
The event extraction module is responsible for extracting the primary event information contained in news text, including latent event types and event arguments. We adopt the ODEE algorithm \cite{LiuHuang-34} to perform this task. The ODEE algorithm and the parameter inference network are illustrated in Fig. \ref{fig2}. We feed the model with clusters of news by date, which have the same related topics. Following the parameter settings in \cite{LiuHuang-34}, we adopt the ADAM optimizer with a learning rate of 0.005 and a moment weight of 0.8. The event slot is chosen by the maximum likelihood after the parameters are estimated, which is formulated as
\begin{equation} 
	\nonumber {p_{\beta^{'},\theta ,\lambda }}\left( {s|e,t} \right) \propto{p_{\beta^{'},\theta ,\lambda }}\left( {s,h,{f^{'}},t} \right) 
	= {p_\theta }\left( {s|t} \right) \cdot {p_\lambda }\left( {h|s} \right) \cdot {p_{{\beta ^{}}}}\left( {{f^{'}}|s} \right)
\end{equation}
We consider utilizing the POS-tags and Staford dependency parser during the event assembling stage to ensure the output is semantically complete. For example, all the words whose POS are Verb or dependency in {advcl, ccomp, rcmod, xcomp} are saved. If a phrase is like "noun.act", "noun.phenomenon", "nun.event" or "noun.attribute", it is also considered. Finally, we save the latent event types and top-5 events for each news cluster, which are the inputs of the LSTM module.

\subsection{ Sentiment Analysis}
The Sentiment analysis module uses Natural Language Toolkit (NLTK) and the Valence Aware Dictionary for sEntiment Reasoner (VADER) to extract the sentiment expressed in news data. Table \ref{table2} illustrates an example of sentiment analysis results. Reference \cite{FeuerriegelHeitzmann-38} forecasted crude oil prices with the news sentiment based on linear regression method and vector autoregressive model, and found that news sentiment significantly enhanced the prediction. Following this idea, we try to use the news sentiment to enhance the predicted mean price. Considering volatility is typically measured by variance, which has no moving direction information. The challenge is that after the GARCH model forecasting the next day's volatility, how can volatility facilitate the mean price prediction directly. News sentiment reflects the game of buying long and selling short in the crude oil market. It can give some directional information and can help mean price prediction by integrating with volatility. Ideally, suppose the sentiment is highly positive, and the volatility is violent. In that case, the higher level the ARIMA fitted mean price can fluctuate to a certain degree towards the positive direction, such as $m$ standard deviation. We use neural networks to character the complex relations. The parameter $m$ can be learned in the following training process. Each news obtains four sentiment scores: negative score, neutral score, positive score, and compound score. We average every day's news sentiment and generate a sentimental vector $se{n_i} = \left[ {ne{g_i},ne{u_i},po{s_i},co{m_i}} \right]$ for the $i-th$ day.

\begin{table}
	\caption{Illustration of news sentiment scores computed by VADER.}
	\label{table2}
	\centering
	\begin{tabular}{ p{20pt} p{240pt} p{30pt} p{30pt}p{30pt}p{30pt}}
		\toprule
		\#& news&  neg&	neu&	pos&	comp \\
		\midrule
		1&	Crude oil on the New York Mercantile Exchange dropped to \$54.40 during early afternoon trading, marking a fifth day of lower oil prices.
		
		&	0.22&	0.78&	0.00&	0.71 \\
		
		2&	Mining shares moved higher on hopes for more Chinese stimulus to boost the country’s economy, with Rio Tinto rising 6.5p to 2873.5p and BHP Billiton 9p better at 1404.5p.
		
		&
		0.00&	0.70&	0.30&	0.86 \\
		
		3&	In 2011, during Barack Obama’s first term in office, the US surpassed Russia as the world’s largest natural gas producer, and in early 2018 it overtook Saudi Arabia as the leading producer of crude oil.&	0.09&	0.85&	0.06&	-0.30 \\
		\bottomrule
	\end{tabular}
	\label{tab2}
\end{table}

\subsection{LSTM Module}
The impact of different news events on the crude oil market has various cycles. For example, the effects of regional military conflicts and import-export policy announcements have different duration. LSTM is adopted to capture the long and short historical information contained in the news events and prices to deal with this problem. We set the hyperparameters based on two aspects: following previous related literature and optimizing the parameters by in-sample data (validation set). According to the input format of LSTM, we trim the news data to 5 news per day and 20 words per event \cite{ChauEsteves-36}. OOV vectors are padded to ensure homogeneous dimensions. The dimensions of event type embeddings are mapped to 100, and dimensions of event argument embeddings are mapped to 200. Assuming that the prices 20 days ago have a negligible effect, we concatenate the latent event type embeddings, event arguments embeddings, and the latest 20 historical prices, acquiring an input vector with 320 dimensions. We take the forward final hidden states as the LSTM module’s forecast price.

The proposed framework, AGESL, is a hybrid framework that integrates the mean price predicted by ARIMA, volatility forecasted by GARCH, sentiment, and the price predicted by LSTM module, by using a fully connected neural network. Each component of the AGESL framework has its unique function. ARIMA module affords a conservative mean price forecast. Sentiment and GARCH modules afford directional volatility, which facilitates mean price moving closer to the actual value at a certain confidence level. The LSTM-based neural network can well capture the nonlinear component of the historical prices and news events. The proposed AGESL integrates the advantages of each module to afford a more reliable crude oil price forecasting.

\section{Experiments}
In this section, we first introduce the statistical performance measurements for event extraction and crude oil price forecast. Then we verify the effectiveness and superiority of the proposed framework AGESL on WTI data and report the experimental results, followed by an analysis and discussion of some observations.

\subsection{ Performance Evaluation}
\textbf{Event Extraction Measurement.} The event extraction statistical evaluation metrics include precision, recall, and F1 score, which are calculated as follows:
\begin{equation} 
	Precision = \frac{{TP}}{{TP + FP}}
\end{equation}
\begin{equation} 
	Recall = \frac{{TP}}{{TP + FN}}
\end{equation}
\begin{equation} 
	\nonumber F1 = \frac{{2*Precision*Recall}}{{Precision + Recall}} = \frac{{2*TP}}{{2*TP + FP + FN}}
\end{equation}
These performance measures provide a brief explanation of the "Confusion Metrics". True positives (TP) and true negatives (TN) are the observations that are correctly predicted. In contrast, false positives (FP) and false negatives (FN) are the values that the actual classes contradict with the predicted classes.

\textbf{Crude Oil Price Forecast Measurement.} The prediction task evaluation metrics include RMSE (Root Mean Square Error), MAPE (Mean Absolute Percentage Error), and DS (Direction Statistics), which have been frequently used in recent research \cite{Abdollahi-3, ZhaoLi-6}. RMSE is a mainstream accuracy indicator describing the forecasting deviation from the actual values:
\begin{equation}
	RMSE = \sqrt {\frac{1}{N}\sum\nolimits_{{\rm{t}} = 1}^N {{{({y_t} - {{\hat y}_t})}^2}} } 
\end{equation}
Considering RMSE is sensitive to data, MAPE is employed to measure the predicted error as a percentage:
\begin{equation}
	MAPE = \frac{1}{N}\sum\nolimits_{{\rm{t}} = 1}^N {\left| {\frac{{{y_t} - {{\hat y}_t}}}{{{y_t}}}} \right|} 
\end{equation}
where ${y_t}$ is the actual value, ${\hat y_t}$ presents the predicted value, and $N$ is the number of total observations in test set. From the view of practical applications, minor forecast errors may not be sufficient to ensure profitable returns because the accurate prediction of direction movement is a more critical criterion. Directional change statistical indicator DS is introduced and formulated as 
\begin{equation}
	DS = \frac{1}{N}\sum\nolimits_{{\rm{t}} = 1}^N {{\alpha _t}} 
\end{equation}
where $\alpha _t$ equals 1 if $({y_{t + 1}} - {y_t}) \cdot ({y_{t + 1}} - {\hat y_t}) \ge 0$ , else equals 0.

\subsection{ Event Extraction Evaluation}
News events are essential predictors for our proposed AGESL framework. We reimplement the experiments executed in \cite{LiuHuang-34} and compare the results with those the authors reported. Table \ref{table3} shows the schema matching evaluation results performed on two data sets: GNBusiness-Test and our news data. The third row and fourth row are results performed on GNBusiness-Test. The third row demonstrates the results reported in \cite{LiuHuang-34}, while the fourth row shows our reimplement results. Unfortunately, we do not get precisely the same results as those the author reported. There are many probable reasons, for example, some tricks or different experimental settings. But, the results do not differ significantly from those reported in \cite{LiuHuang-34}. It shows that the ODEE algorithm could afford favorable event extraction results.

To check the generalization, we annotate 14 batches of news clusters on our news data following the same settings and then perform the event extraction task. We report the schema matching results performed on our data set in the fifth row in Table \ref{table3}. One advantage of open-domain event extraction is finding the unseen events, so we expect the model to have higher recall values. Overall, the recall 59.6\% and F1 score show that ODEE still performs well on our news data.

In summary, Table \ref{table3} demonstrates the results performed on two different data sets, and the purpose is to check ODEE's schema matching generalization. As to the open-domain event extraction task, except for the schema, the arguments and roles are the key points of the events, the overall performance shows ODEE still performs well on our news data.

\begin{table}
	\caption{Schema matching evaluation.}
	\label{table3}
	\centering
	\begin{tabular}{p{150pt} p{70pt} p{70pt} p{70pt} }
		\toprule
		\multirow{2}{*}{News dataset}& \multicolumn{3}{c}{Schema Matching (\%)} \\
		\cline{2-4} 
		&  Precision&	Recall& F1  \\   
		\midrule
		GNBusiness-Test \cite{LiuHuang-34}&		43.4&	58.3&	49.8 \\
		GNBusiness-Test (we reimplement)&		41.8&	56.7&	48.1 \\
		Our news data&	39.1&	59.6&	47.2 \\
		\bottomrule
	\end{tabular}
	\label{tab3}
\end{table}

\subsection{ Crude Oil Price Forecast Evaluation}
Six mainstream forecasting models are selected as benchmarks to demonstrate the effectiveness and superiority of the proposed framework AGESL.
\begin{itemize}
	\item \textbf{ARIMA} refers to the ARIMA module in AGESL, and the parameters settings such as orders are the same.
	\item \textbf{SVM} was initially developed for classification tasks but has been extended to solve nonlinear regression and time series prediction problems \cite{XieYu-5}. The parameter settings follow \cite{WuWang-4}.
	\item \textbf{LSTM} takes only price features as input.
	\item \textbf{LSTM-Sent} takes prices and news sentiment as input.
	\item \textbf{LSTM-Event} takes prices and news events as input. 
	\item \textbf{ARIMA-GARCH-Sent} is a hybrid method that integrates the ARIMA module, GARCH module, and news sentiment. The parameter settings are compatible with AGESL.
\end{itemize}
We do our best to optimize the parameters of the listed models by in-sample data (validation set), then conduct experiments on out-of-sample data. We repeat every model 10 times and report the average results in Table \ref{table4}. 
From Table \ref{table4}, some findings can be summarized as follows:

\begin{table}
	\caption{Crude oil price forecast results on test data set.}
	\label{table4}
	\centering
	\begin{tabular}{ p{150pt} p{70pt} p{70pt} p{70pt} }
		\toprule
		Methods&	RMSE&	MAPE&	DS \\
		\midrule
		ARIMA&	1.4846&	0.0256&	0.5180 \\
		SVM&	1.3056&	0.0224&	0.5967 \\
		LSTM&	1.2823&	0.0218&	0.6147 \\
		LSTM-Sent&	1.1612&	0.0199&	0.6459 \\
		LSTM-Event&	1.1003&	0.0184&	0.6574 \\
		ARIMA-GARCH-Sent&	1.2736&	0.0213&	0.6393 \\
		\textbf{AGESL}&	\textbf{1.0825}&	\textbf{0.0179}&	\textbf{0.6885} \\	
		\bottomrule
	\end{tabular}
	\label{tab4}
\end{table}

Firstly, the models (LSTM-Sent, ARIMA-GARCH-Sent, LSTM-Event, and AGESL) that take multi-channel information as input outperform those (ARIMA, SVM, and LSTM) with single-channel information. The reason is that many factors affect crude oil prices, even some of which are hard to quantify. Single-channel information (e.g., historical prices) is insufficient to accurately predict the future oil price. Recent research on crude oil price forecasting has considered the qualitative effectiveness of political events, regional conflicts, and policies of oil-exporting countries, and so on. Online news, especially collected from formal news portals (e.g., The New York Times, The Guard), contains extensively relative information, and the contents are less noisy and more objective. Big-data and NLP technologies provide measures to quantify these influencing factors. This study extracts two kinds of features from news text, including news events and news sentiment. News events and sentiment are relatively complementary because the former reflect major events that may influence the crude oil market. The latter may affect the buyers' and sellers' potential behaviors. The multi-channel models that fully utilize these features averagely achieve better accuracy than single-channel models in terms of lift 14.96\% on RMSE, 16.72\% on MAPE, and 14.10\% on DS, respectively.

Secondly, as to single-channel input models (including ARIMA, SVM, and LSTM), it can be found that LSTM outperforms better than ARIMA and SVM. There are three possible reasons. First, ARIMA is a linear method that is insufficient to capture the complex nonlinear characters. Second, SVM and LSTM are nonlinear models and thus obtain better results than ARIMA. Third, LSTM is good at processing time-series or sequence-to-sequence data because LSTM cells have a memory mechanism that stores previous timestep information.

Thirdly, comparing the LSTM and LSTM-Sent model, it can be found that news sentiment can significantly enhance the prediction performance and gives a 3.1\% gain in DS. Comparing the LSTM-Sent and LSTM-Event, we find that although news sentiment and news events are both features extracted from news text, LSTM-Event gives lower forecast errors and a 1.2\% gain in DS. It demonstrates that news events contain more helpful information. News sentiment and event extraction approaches we adopted in this study are all unsupervised learning methods. The differences between them are that higher dimensional continuous latent event type embeddings have more vigorous representation, and event arguments word embeddings contain external knowledge. Comparing the ARIMA-GARCH-Sent and AGESL, it demonstrates that the news events and sentiment are relatively complementary information and can be combined to enhance forecast performance.

Finally, the proposed AGESL approach outperforms all other benchmarks. It achieves the highest DS, the lowest RMSE, and MAPE among all models. Compared with all the other benchmark models, it averagely achieves better accuracy in terms of lift 14.62\% on RMSE, 17.00\% on MAPE, and 12.50\% on DS. Even Compared with the suboptimal LSTM-Event, the RMSE, MAPE, and DS of AGESL obtain a lift of 1.62\%, 2.72\%, and 4.73\%, respectively. It demonstrates that a hybrid framework integrating the advantages of its subcomponents is more capable of crude oil price forecasting.

\subsection{Discussion}
To better analyze the proposed AGESL, we will further discuss some characteristics for forecasting crude oil prices, including the effectiveness of event extraction, result visualization, and significance test of the difference between AGESL and benchmarks.

\subsubsection{The Effectiveness of Event Extraction}
Oil prices are not only driven by economic variables but also affected by critical events, such as political events, military conflicts, severe climate abnormalities, and even significant accidents \cite{FanLiang-49}. The news about crude oil expressed online and in newspapers partially represents the crude oil industry events. Much research has studied the news effect on crude oil price and utilized the news to facilitate crude oil price forecasting. However, much existing literature mainly uses the news headlines directly \cite{WuWang-8}, or uses features generated by topic models  \cite{ZhaoLiu-40, LiShang-42}. Different from previous work, we use Event Extraction, an advanced form of Information Extraction, to extract open domain events and use the results to facilitate crude oil price prediction. From Table \ref{table4}, we can find that the models with the Event Extraction module (LSTM-Event and AGESL) are the top two optimal models by any evaluation metric from RMSE, MAPE, and DS. Especially compared with the LSTM model, the LSTM-Event model gets lower forecast errors and higher direction accuracy, for example, 4.3\% gain in DS. The results demonstrate the effectiveness and superiority of event extraction. The probable reason is that event extraction not only extracts latent event types, but also extracts arguments which maintain the primary abstract information and reduce the irrelevant information.

\subsubsection{Result Visualization}
\begin{figure*}[htbp]
	\centering
	\includegraphics[width= 6.3 in]{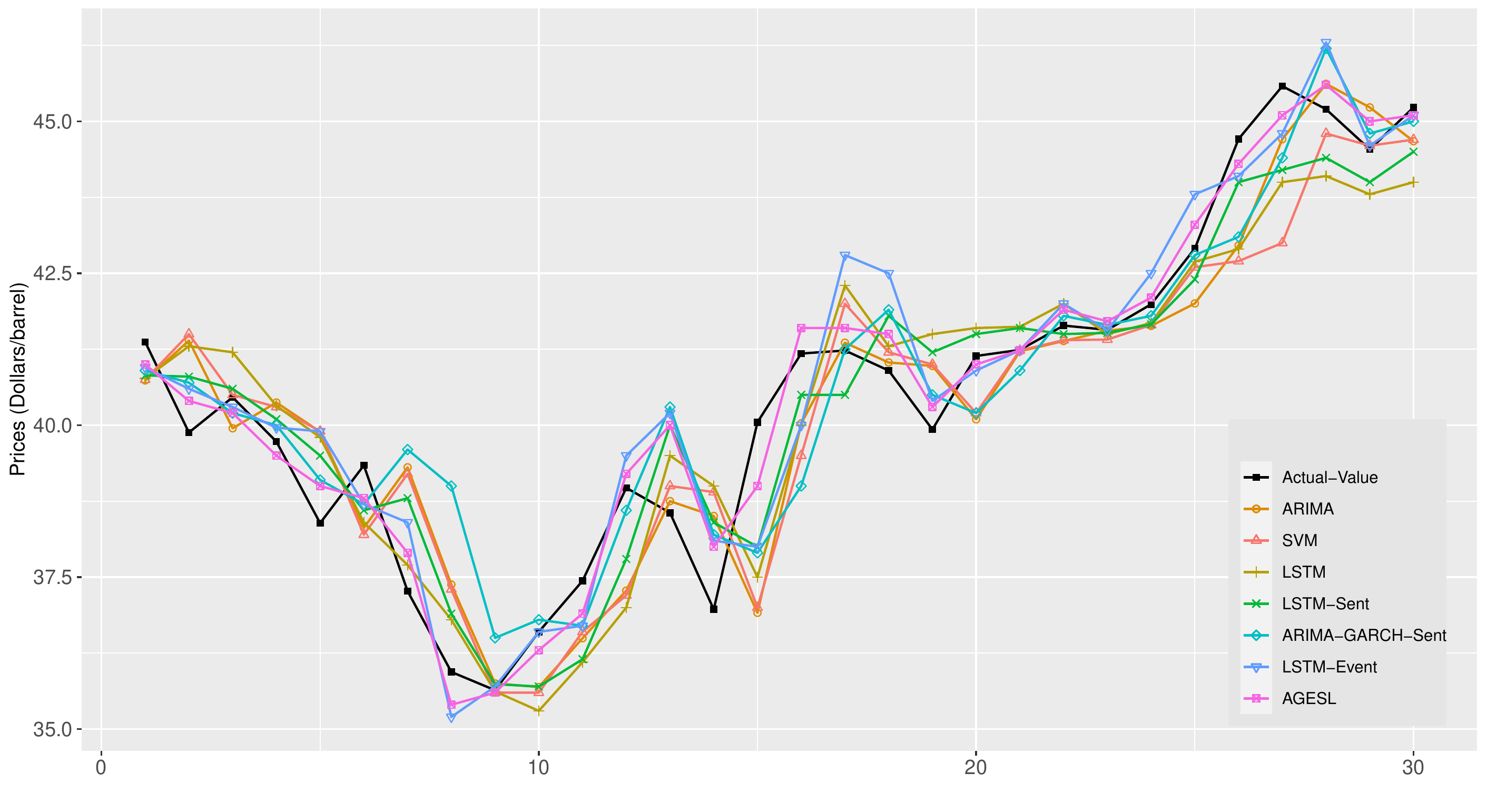}	
	\caption{Actual values and predicted values for sampling test data. Using the full testing data set will result in lines in the figure hard to distinguish. Here, we randomly sample a period including 30 successive trading days and plot the actual values along with the corresponding predicted prices.}
	\label{fig5}	
\end{figure*}

To visually compare the performance of each model, we randomly sample a period including 30 successive trading days and plot the corresponding predicted prices in Fig. \ref{fig5}. We sample the data because using the full testing data set will result in lines in the figure hard to distinguish. From Fig. \ref{fig5}, we can see that the AGESL line is closer to the actual-value line than other benchmark models.

\subsubsection{Significance Test of Difference}
To test whether the obtained values of these criteria are statistically equal, researchers can rely on the methodology proposed by Diebold and Mariano (1995) (DM) test \cite{Franses-51, Diebold-52}. The Diebold-Mariano (DM) test was intended for comparing forecasts; it has been, and remains, helpful in that regard \cite{XuAamir-44, GaoAamir-45}. To check the superiority of the proposed model AGESL, we conduct the DM test to check its statistical significance. The DM statistics with their p values are shown in Table \ref{table5}. The proposed model AGESL statistically outperforms ARIMA, SVM, LSTM, LSTM-Sent, LSTM-Event, ARIMA-GARCH-Sent models with their respective p values below 0.01. Thus the predictive ability of the proposed AGESL framework is verified with statistical evidence.

From the above discussion and analysis, we can conclude that with the benefits of the Event Extraction module, the proposed framework AGESL significantly outperforms all other models listed in this paper in terms of RMSE, MAPE, DS, and DM test.

\begin{table*}
	\caption{The Diebold-Mariano (DM) test results of WTI crude oil prices.}
	\label{table5}
	\setlength{\tabcolsep}{3pt}
	\begin{tabular}{p{40pt} p{50pt} p{50pt} p{50pt} p{60pt}  p{70pt} p{100pt}}
		\hline
		\makecell[c] {Model}&	\makecell[c] {ARIMA}&	\makecell[c] {SVM}&	\makecell[c] {LSTM}&	\makecell[c] {LSTM-Sent}&	\makecell[c] {LSEM-Event}&	\makecell[c] {ARIMA-GARCH-Sent} \\
		\hline
		\makecell[c] {AGESL}&	{\makecell[c]{$-$17.87 $\ast$ \\(<0.01)}}&	{\makecell[c]{$-$14.70 $\ast$ \\(<0.01)}}&	{\makecell[c]{$-$18.22 $\ast$ \\(<0.01)}}&	{\makecell[c]{$-$10.19 $\ast$ \\(<0.01)}}&	{\makecell[c]{$-$8.56 $\ast$ \\(<0.01)}}&	{\makecell[c]{$-$11.67 $\ast$ \\(<0.01)}} \\
		\hline
		\multicolumn{7}{p{251pt}}{"$\ast$" significant at 1\%. }\\
	\end{tabular}
	\label{tab5}
\end{table*}

\section{Conclusion}
Crude oil prices may be affected by some factors that are hard to quantify, such as political events, regional conflicts, and policies of oil-exporting countries, which are frequently reported in online news. In this paper, we propose a new hybrid framework, AGESL, for predicting crude oil prices whose focus is on capturing these features to enhance the prediction accuracy. For this purpose, we utilize an open-domain Event Extraction algorithm and other NLP technologies to extract news events (e.g., latent event types and news arguments) and news sentiment from news text. The proposed AGESL integrates the strength of linear and nonlinear modules. The proposed AGESL outperforms the other benchmark models in the empirical study, including the single-channel input models (ARIMA, SVM, LSTM) and multi-channel input models (LSTM-Sent, ARIMA-GARCH-Sent, LSTM-Event).

Despite the promising capability of the proposed AGESL framework, it may be further developed from the following perspectives. First, from the export-import view, the world's major economies consume the vast majority of crude oil, so more economic or financial exogenous variables, such as economic growth, industrial added value, or financial indices for these economies need to be considered into the framework. Second, from the view of practical applications, market risk should be taken into account to minimize a market trader's potential loss. For example, the VaR method could help the proposed AGESL report a forecast value and the corresponding risk at a certain confidence level.

\bibliographystyle{unsrt}  
\bibliography{ReferFromNotEp}

\end{document}